\documentclass{article}

\usepackage{PRIMEarxiv}

\usepackage[utf8]{inputenc} 
\usepackage[T1]{fontenc}    
\usepackage{hyperref}       
\usepackage{url}            
\usepackage{booktabs}       
\usepackage{amsfonts}       
\usepackage{nicefrac}       
\usepackage{microtype}      
\usepackage{lipsum}
\usepackage{apacite}
\usepackage{fancyhdr}       
\usepackage{graphicx}       
\usepackage{algorithm}
\usepackage{algpseudocode}
\usepackage{amsmath}
\usepackage{multirow}
\usepackage{adjustbox}
\usepackage{natbib}
\usepackage{xcolor}
\graphicspath{{media/}}     

\title{Interpreting Public Sentiment in Diplomacy Events: A Counterfactual Analysis Framework Using Large Language Models 
}

\author{
  Leyi Ouyang \\
  South China Agricultural University Zhujiang College \\
  \texttt{Leyiouyang@163.com} \\
}

\begin{document}
\maketitle

\begin{abstract}
Diplomatic events consistently prompt widespread public discussion and debate.
Public sentiment plays a critical role in diplomacy, as a good sentiment provides vital support for policy implementation, helps resolve international issues, and shapes a nation's international image. Traditional methods for gauging public sentiment, such as large-scale surveys or manual content analysis of media, are typically time-consuming, labor-intensive, and lack the capacity for forward-looking analysis. We propose a novel framework that identifies specific modifications for diplomatic event narratives to shift public sentiment from negative to neutral or positive. First, we train a language model to predict public reaction towards diplomatic events. To this end, we construct a dataset comprising descriptions of diplomatic events and their associated public discussions. Second, guided by communication theories and in collaboration with domain experts, we predetermined several textual features for modification, ensuring that any alterations changed the event's narrative framing while preserving its core facts.
We develop a counterfactual generation algorithm that employs a large language model to systematically produce modified versions of an original text. The results show that this framework successfully shifted public sentiment to a more favorable state with a 70\% success rate. 
This framework can therefore serve as a practical tool for diplomats, policymakers, and communication specialists, offering data-driven insights on how to frame diplomatic initiatives or report on events to foster a more desirable public sentiment.

\end{abstract}


\section{Introduction}
In the realm of international relations, diplomatic events serve as pivotal tools through which states and international actors pursue their objectives, manage relationships, and address global challenges \citep{leguey2017global,ociepka2018public}. 
These events include, for example, negotiations between nations, multilateral summits addressing global issues, formal public declarations stating policy positions, and the strategic implementation of various foreign policy tools \citep{ghosh2020international}. Throughout diplomatic events, the public actively reacts, discusses, and comments, shaping a sentiment \citep{yiu2022public}. Positive public sentiment can be useful in promoting specific diplomatic initiatives \citep{rhee2024perceived, huh2024setting, yiu2022public}. In addition, favorable public sentiment can contribute to the projection of a compelling national vision, thus enhancing broader domestic and international support. Public sentiment can also affect and guide how diplomatic events develop \citep{goldsmith2009spinning}. Strongly positive public sentiment often encourages leaders to pursue more ambitious diplomatic goals and take bolder actions. In contrast, significant public disapproval typically leads to increased caution, prompting leaders to limit their goals or avoid controversial policies \citep{hartig2017deterrence,holsti2009public,barston2019modern}.

In light of its impact, effectively gauging public sentiment towards diplomatic events is a key pursuit for both international actors and researchers. Traditionally, efforts to understand public opinion have relied on established methods such as direct polling through surveys, qualitative analysis of media coverage, and expert assessments \citep{ferguson2000researching}. Although these approaches offer a useful assessment of public sentiment, they are typically time-consuming, labor-intensive, and lack the capacity for forward-looking analysis.
Therefore, we aim to explore how to efficiently measure public sentiment in response to diplomatic events. 

Numerous studies attempted to use machine learning approaches to model the connection between real-world events and the dynamics of public sentiment, to try to improve the prediction of public sentiment faster and more accurately \citep{cederman2023computational, ouyang2025can, cao2024ecc}. For understanding the textual data, deep learning architectures such as Long Short-Term Memory (LSTM) networks are specifically designed to address the challenges of learning dependencies in text \citep{sherstinsky2020fundamentals}. Studies have demonstrated its ability to understand comments on social media (such as Weibo and Twitter), product reviews, and political discourse \citep{shi2017hierarchical,yang2021cross,yan2024constructing}. However, LSTMs can still struggle with very long-range contextual dependencies \citep{challagundla2024neural}. Consequently, these models are less powerful for comprehensively capturing the overall and intricate relationships in the diplomatic events. 

Transformer-based models, such as BERT (Bidirectional Encoder Representations from Transformers), GPT, mark a significant advancement in understanding and capturing key information for long text. Additionally, its pre-training on vast text corpora allows BERT to develop a meaningful representation of words \citep{yildirim2024mastering}. Consequently, BERT-based models typically have stronger performance for general text understanding tasks \citep{zhou2024comprehensive}. However, a primary challenge lies in applying such a model trained on general text materials to the specialized context of the diplomatic domain. The language in this area is often characterised by specific terminologies, ambiguity, indirectness, irony, sarcasm, and strategic hedging. Therefore, the general BERT models may fail to grasp the intended meaning of diplomatic texts fully, potentially leading to inaccurate or even misleading interpretations \citep{yadav2024contextual}. Additionally, BERT models are considered as a ``black box'', posing a considerable challenge for interpreting the prediction \citep{hassija2024interpreting, ahmed2023deep, wang2023sparsity}. However, understanding which features of the event or linguistic patterns influenced the sentiment prediction is important.

In this study, we propose a novel framework that integrates both the prediction and change of public sentiment.
There are two main components in this framework. The first is a public sentiment proxy. We fine-tune a BERT model to predict the public sentiment for diplomatic events. To fine-tune a model for the specific domain of diplomacy, we construct a dataset comprising descriptions of diplomatic events and their associated public discussions. The second component of our framework uses counterfactual analysis to systematically modify the original event texts from several key diplomatic dimensions—such as the participants involved, the event process, and communication type—to understand how specific changes can shift the public sentiment. Our experimental results demonstrate that for events initially classified as `Negative,' our algorithm identified a sequence of alterations that shifted their sentiment to the `Neutral' or `Positive' with a 70\% success rate. We also examine how sentiment for the same diplomatic events varied across different user communities. We find that different groups often react quite differently to the same event.

This framework offers significant practical applications for policymakers, diplomats, and communication specialists. By not only predicting public sentiment but also identifying its specific drivers, our approach serves as a strategic tool for risk management and public relations. It can be used to simulate and anticipate public reactions to planned diplomatic initiatives, allowing officials to refine their communication strategies proactively. For events that have already occurred, the framework can help diagnose why public sentiment is negative and identify potential narrative adjustments for crisis communication or public opinion guidance, providing a data-driven method for fostering a more favorable public reception.

The remainder of this paper is organised as follows. Section \ref{sec:literaturereview} presents research on public sentiment in diplomatic contexts. This section also reports key communication theories that support our proposed framework's design. Section \ref{sec: Counterfactual framework} details our counterfactual analysis framework. Section \ref{sec:experimentsandresultsanalysis} describes the experimental setup, presents the results, and analyzes them. Section \ref{sec:conclusion} concludes the paper and discusses the implications of our findings.

\section{Literature Review}
\label{sec:literaturereview}
\subsection{Methodologies of measuring public opinion}
Traditionally, the common approach is public opinion polling or surveys, where researchers ask a representative sample of the population direct questions about their views on a specific foreign policy or event \citep{berinsky2017measuring}. Researchers also analyze the content of newspaper articles and editorials or conduct focus groups \citep{moretti2011standardized}. Additionally, some studies analyze public behavior patterns for different regions \citep{hong2012public}.
While these approaches provide valuable insights, they are generally labor-intensive, limited in scale, and lack the capacity for the forward-looking analysis needed to anticipate public reactions.

Many early studies used machine learning to predict public opinion. This approach relies on researchers using their domain knowledge to predefine and extract a set of measurable variables from various media, which then serve as structured input for a predictive model \citep{weiss2010text}.
For example, to predict public sentiment from a news article, a researcher might extract features such as:
(1) Keyword Frequency: The number of times words like ``crisis," ``breakthrough," or ``tension" appear \citep{de2013discourse}. (2) Sentiment Lexicon Score: A score calculated by counting the positive and negative words in the text based on a predefined dictionary \citep{taboada2011lexicon}. (3) Text Statistics: Basic metrics like the total word count, average sentence length, or the percentage of uppercase words \citep{li2020textual}. (4) Readability Score: A metric like the Flesch-Kincaid index to measure the complexity of the language \citep{solnyshkina2017evaluating}. (5) Vocal tone, pitch, or speaking rate in speeches or news broadcasts to quantify emotional intensity or sentiment\citep{dietrich2019pitch}.
These hand-crafted features transform the raw text into a numerical format that a machine learning model can use to learn patterns and make predictions.
Beyond these explicit, hand-crafted variables, text contains a wealth of latent features—deep semantic patterns and contextual relationships—that are highly valuable for making accurate predictions.

To capture these latent semantic patterns in language, researchers use NLP techniques to analyze diplomatic events. \cite{zhou2022policy} uses to classify policy documents into different types of events. \cite{lin2021techniques}  predict international conflicts. Recent work by \cite{wessel2023improving} leveraged transformer models to detect framing biases in news coverage of diplomatic summits.  \cite{cao2024catmemo} fine-tune a language model for a specific domain and task. While these NLP techniques demonstrate a strong capacity for prediction and classification, a significant limitation is that they often do not provide actionable insights. That is, they can predict outcomes like public sentiment or international conflict, but do not inherently recommend how one might modify diplomatic communication to achieve a more desirable public response. This challenge is compounded by the ``black-box" nature of many sophisticated models. It remains difficult to interpret precisely which event features or linguistic patterns the model used to make its decision. This lack of interpretability makes it hard to derive actionable guidance from the predictions alone. 

Much of this computational analysis relies on social media data as a proxy for real-time public opinion. Platforms like Twitter and Reddit provide vast, real-time streams of public expression where users actively discuss and comment on diplomatic events as they unfold. This rich, conversational data allows researchers to capture immediate public reactions on an unprecedented scale, offering a dynamic alternative to the more static insights provided by traditional methods.


\subsection{Narrative framing and its impact on public opinion}
The narrative used to describe an event is a powerful force in shaping public reaction. Two reports describing the exact same event can generate vastly different public sentiments based purely on how they are framed \citep{entman2009projections}.
By strategically choosing certain words, emphasizing particular details, or highlighting specific actors, a narrative can guide the audience's interpretation and emotional response \citep{dahlstrom2014using}. The story told about an event is often as influential as the event itself. For example, \cite{baartman2023impact} demonstrated how narratives about diplomatic initiatives, such as climate accords or trade deals, are meticulously crafted to evoke positive sentiment. 

Framing theory explains how the public meaning of a diplomatic event is shaped by selectively emphasizing certain aspects of it \citep{entman1993framing}. Framing involves making some narrative elements more salient to define problems, diagnose causes, or evaluate outcomes. In diplomacy, governments and the media use frames to attribute responsibility or mobilize public support \citep{iyengar1990shortcuts, scheufele1999framing}. For instance, \cite{chong2007framing} showed in their experimental work that changing a frame’s emotional tone can significantly shift public perception of international agreements. 

Beyond framing, Agenda-Setting Theory explains how media coverage influences public opinion by making certain diplomatic events more salient than others \citep{mccombs2020setting}. By controlling which events receive attention, the media not only tells the public what to think, but also what to think about, thereby elevating the importance of an issue. Closely related is the priming effect, which suggests that this salience primes the audience to evaluate leaders based on their handling of those specific, high-profile issues \citep{althaus2006priming}. For example, extensive coverage of a climate summit may lead the public to judge a leader's entire foreign policy based on their performance on climate change.

How an audience processes these messages is also explained by Narrative Persuasion Theory \citep{bilandzic2013narrative}. This theory suggests that when people become absorbed in a story, their cognitive resistance to persuasion is lowered. A compelling narrative about a diplomatic ``breakthrough" can therefore be more influential than a list of policy facts, as it engages the audience emotionally and makes them less likely to counter-argue. Together, these theories illustrate a clear pathway: the media sets the agenda of important events (Agenda-Setting), primes the criteria for evaluation (Priming), shapes the narrative of those events (Framing), and can make the message more persuasive by engaging the audience in a story (Narrative Persuasion).

\subsection{Explainable AI and Counterfactual Reasoning}
While powerful models like BERT can accurately predict public sentiment, their ``black-box" nature often makes it difficult to understand why they arrive at a specific conclusion. This has led to the rise of Explainable AI (XAI), a field dedicated to making complex models more transparent and interpretable.

A key technique within XAI is counterfactual reasoning, which probes a model's logic by asking ``what if" questions. By systematically changing parts of an input—such as the features of a diplomatic event narrative—and observing how the model's prediction changes, researchers can infer which elements are the most influential drivers of the outcome. This approach effectively opens the black box, providing a clear, cause-and-effect method for understanding a model's decision-making process.

\section{Counterfactual Framework}
\label{sec: Counterfactual framework}
\begin{figure*}
    \centering
    \includegraphics[width=1\linewidth]{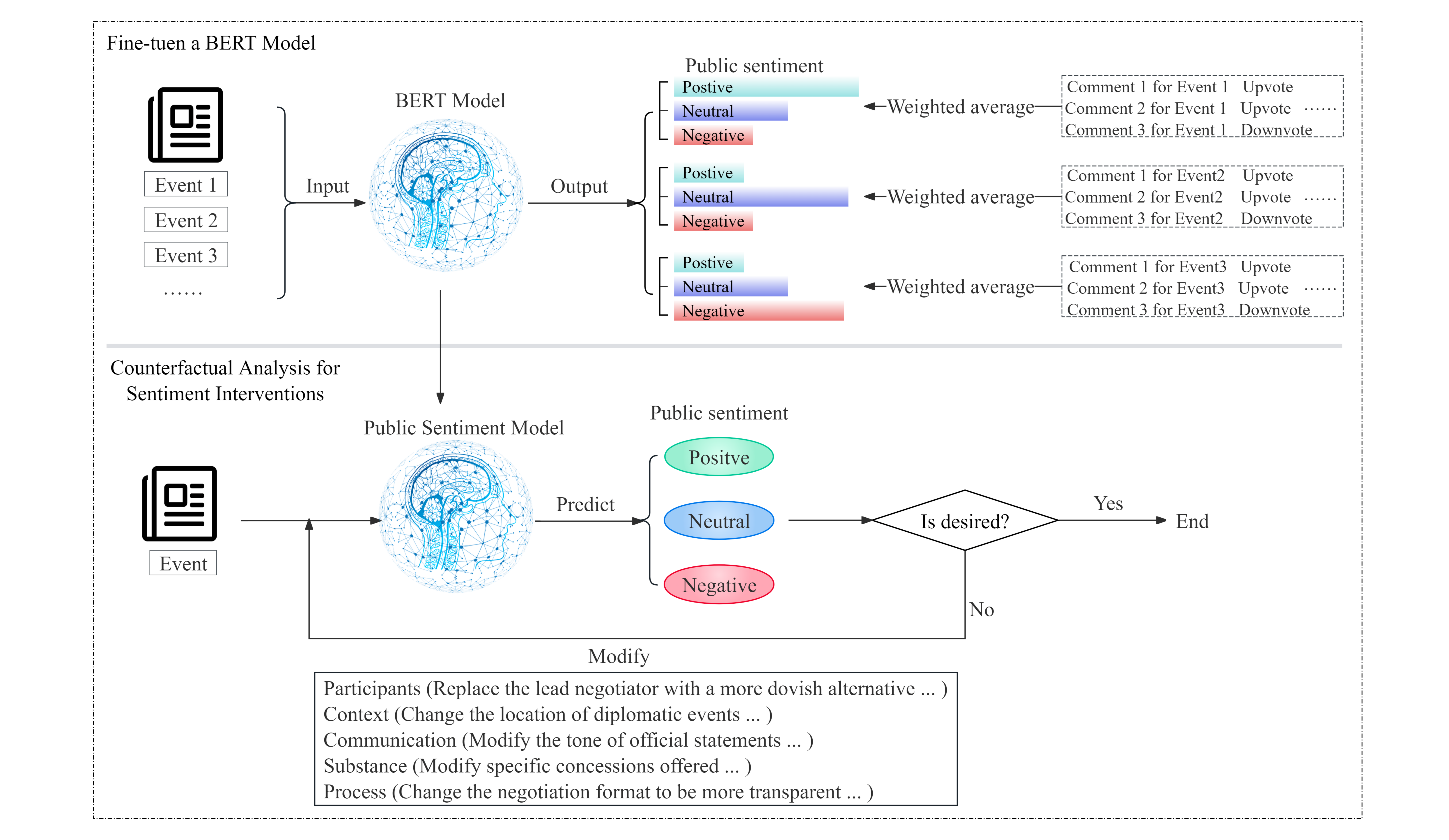}
    \caption{Counterfactual Framework}
    \label{fig:flowchart}
\end{figure*}

This section details the counterfactual framework, as illustrated in Figure \ref{fig:flowchart}, which is designed to predict and interpret public sentiment on diplomatic events. We begin by explaining the process of fine-tuning a language model to accurately predict public opinion, allowing it to serve as a reliable proxy for public sentiment. We then introduce the counterfactual algorithm used to systematically alter the event narratives for analyzing how these changes impact the model's predictions.

\subsection{A Fine-tune BERT Model for Public Sentiment}
For quantifying public sentiment towards diplomatic events, we selected Reddit as its primary data source due to several distinct advantages. As a widely utilised platform, Reddit facilitates the sharing of news articles regarding diplomatic occurrences and hosts extensive user discussions, offering a rich repository of public reactions directly linked to these specific events \citep{chancellor2020methods}. These discussions provide valuable insights into how diplomatic events are framed, debated, and received by various online user communities. Furthermore, the platform's pseudo-anonymous nature can encourage more candid and unfiltered expressions of opinion compared to other social media environments \citep{beian2021increasing}. Crucially, Reddit’s interactive voting system, characterised by `upvotes' and `downvotes' on posts and comments, offers an additional layer of quantifiable data reflecting community agreement or disagreement with specific content or viewpoints. Moreover, Reddit's architecture, featuring diverse, topic-specific communities known as ``subreddits" (such as r/worldnews, r/geopolitics, or various national subreddits), enables a granular analysis of sentiment, allowing for the examination of opinions across different online publics and potentially reflecting varied ideological spectrums.

To quantify public sentiment for diplomatic events, our approach begins with collecting posts where users reference diplomatic events, along with their associated comment threads. Each comment is then analysed using a BERT model specialised for sentiment classification (bert-base-multilingual-uncased-sentiment). This model was fine-tuned on a large corpus of Chinese and English texts, including news articles, social media posts, and official statements related to international relations \citep{jia2020entity} \citep{devlin2019bert}.
This model outputs probabilities for three classes: positive, neutral, and negative, with the highest probability indicating the comment's primary sentiment category. To create a nuanced metric reflecting both positive and negative intensity, we calculate a `compounding sentiment score' for each comment by subtracting its predicted negative probability from its positive probability. Separately, to account for community validation, a `comment score' is derived from Reddit's voting system (upvotes minus downvotes). This comment score is subsequently used to determine a weight for each comment's sentiment. Finally, an aggregated public sentiment score for each original Reddit post is computed by taking the weighted average of these compounding sentiment scores from all comments under that post. At this point, each diplomatic event in our dataset is paired with a corresponding quantitative public sentiment score.

Next, we use these event-sentiment pairs to fine-tune a language model to learn the relationship between the descriptive features of a diplomatic event and the public reaction it generates. The goal is to create a robust model that, given a description of any new diplomatic event, can accurately predict the resulting public sentiment. Since different language models possess unique architectural characteristics and advantages, we compared the performance of several models on this prediction task to identify the most effective approach, as presented in 
Table \ref{tab:Accuracy Comparison of Language Models for Learning to Predict Public Sentiment from Diplomatic Events}. The evaluated models include deep learning approaches like LSTM and CNN, alongside Transformer-based models such as BERT, Sentence-BERT, and Llama 1.1 B. We use two distinct inputs of narrative of diplomatic events: article titles alone and titles combined with their full text.

A consistent observation across all models is that incorporating the full article text alongside the title invariably improves predictive accuracy, underscoring that richer textual context provides more salient information for sentiment determination. Traditional deep learning models, LSTM and CNN, while also benefiting from the added text, demonstrated lower overall accuracy compared to most of the Transformer-based architectures. This performance difference likely stems from the more advanced structural features, such as attention mechanisms and larger model sizes, characteristic of the Transformer models.

Among the models, BERT achieved the highest accuracy overall, highlighting its robustness for this sentiment prediction task. Sentence-BERT exhibited a notable pattern, displaying the most substantial increase in accuracy when provided with the full text in addition to the title. This significant gain might be attributed to Sentence-BERT's typical pre-training on tasks like semantic similarity or question answering, thereby enhancing its ability to understand context and causal relationships. Finally, Llama 1.1B, despite its status as a large language model, showed surprisingly modest performance. A plausible reason for this is that its general-purpose pre-training could conflict with the specialised terminology and nuances of diplomatic discourse. Insufficient domain-specific fine-tuning data for our particular task might then hinder its predictive accuracy in this specific context, as its general knowledge could be ``mixed up" by domain-specific material without enough targeted training.

\begin{table}[htbp]
  \centering
  \caption{Accuracy Comparison of Language Models for Learning to Predict Public Sentiment from Diplomatic Events}
    \begin{tabular}{lrrrrr}
    \toprule
          & \multicolumn{1}{l}{LSTM} & \multicolumn{1}{l}{CNN} & \multicolumn{1}{l}{BERT} & \multicolumn{1}{l}{Sentence BERT} & \multicolumn{1}{l}{Llama-1.1B} \\
    \midrule
    Title & 0.56  & 0.55  & 0.60  & 0.48  & 0.55 \\
    Title+Text & 0.57  & 0.59  & 0.70  & 0.62  & 0.58 \\
    \bottomrule
    \end{tabular}%
  \label{tab:Accuracy Comparison of Language Models for Learning to Predict Public Sentiment from Diplomatic Events}%
\end{table}%

In this study, we fine-tuned a BERT model for predicting public sentiment. This model achieved a decent overall accuracy of 70\%. However, it can potentially be improved if we take a closer examination of its performance, detailed in the classification report (Table \ref{tab:Classification Report of Fine-tune BERT Model}). The model demonstrates strong predictive capabilities for the `Negative' sentiment category, achieving robust precision, recall, and F1-scores. The predictions for the `Positive' class also indicate a reasonable level of performance. However, the model's overall accuracy appears to be primarily constrained by its significantly weaker performance on the `Neutral' sentiment class. This difficulty in accurately identifying neutral sentiment is likely attributable to the insufficient volume of training data available for this specific category, which was notably less than for other classes, pointing to a data imbalance issue that impacted model learning for this particular sentiment. Therefore, it is anticipated that augmenting the training dataset with a larger and more representative sample of neutral instances could significantly enhance the model's predictive accuracy for this class and consequently improve its overall performance.

\begin{table}[htbp]
  \centering
  \caption{Classification Report of Fine-tune BERT Model}
    \begin{tabular}{lrrrr}
    \toprule
          & \multicolumn{1}{l}{Precision} & \multicolumn{1}{l}{Recall} & \multicolumn{1}{l}{F1-score} & \multicolumn{1}{l}{support} \\
    \midrule
    Negative & 0.75  & 0.88  & 0.81  & 849 \\
    Neutral & 0.4   & 0.1   & 0.16  & 274 \\
    Positive & 0.48  & 0.48  & 0.48  & 288 \\
    \bottomrule
    \end{tabular}%
  \label{tab:Classification Report of Fine-tune BERT Model}%
\end{table}%

\subsection{Iterative Counterfactual Generation Algorithm}
Counterfactuals are statements or scenarios that explore what might have occurred if certain past conditions or events had been different. This process involves altering a specific feature.

To generate meaningful counterfactual scenarios, we first defined several key categories of diplomatic event features (see Table \ref{tab:counterfactual_categories}), based on the communication theories. (1) Participants: alter who is involved (e.g., replacing a negotiator). The perceived character of the actors influences message reception \citep{cohen2018defining}. (2) Process: modify procedural aspects (e.g., making negotiations more transparent). This draws from theories of procedural justice, which posit that fair processes can enhance legitimacy and public trust, independent of the outcome \citep{tyler2003procedural}. (3) Communication: change the tone or framing of statements. Modifying the tone engages with a frame's emotional attitude, while reframing key issues alters how a problem is defined and understood by the public \citep{kaufman2003frames}. (4) Substance: alter the core concessions or objectives of an event. This allows for an analysis of how the public responds to different policy proposals and persuasive arguments. (5) Context: modify the situational or symbolic backdrop of an event (e.g., its location). This also relates to Framing Theory, as contextual details can prime audiences and change the salience of different event aspects.

Within each category, specific types of modifications were pre-established; for instance, under ``participants", modifications included replacing lead negotiators or changing stakeholder inclusion, while "communication" involved altering the tone of statements or reframing issues (more details are in Table \ref{tab:counterfactual_categories}). 
The actual generation of counterfactual texts was performed by prompting a large language model. This involved a two-part prompt structure: a system prompt instructed the model to act as an expert in diplomatic communication, tasked with modifying a given text according to a specified category and modification type, while maintaining core facts and creating a plausible alternative of similar length. The human prompt then provided the original news article text and reiterated the specific modification to be applied (e.g., ``change the participants aspect by replacing the lead negotiator with a more dovish alternative").

\begin{table*}[htbp]
  \centering
  \caption{Counterfactual Categories for Diplomatic Analysis}
    \begin{tabular}{p{4cm}p{11cm}}
    \toprule
    Counterfactual Category & Example Modification \\
    \midrule
    Participants & Replace the lead negotiator with a more dovish alternative\newline{}Include additional stakeholders in the negotiations\newline{}Exclude certain parties from the talks \\
    Process & Change the negotiation format to be more transparent \newline{}Modify the timing of diplomatic initiatives \newline{}Alter the use of coercive measures (e.g., sanctions, incentives) \\
    Communication & Modify the tone of official statements (e.g., more conciliatory, more assertive)\newline{}Change the level of publicity for negotiations (e.g., public vs. private talks)\newline{}Reframe key issues in different terms \\
    Substance & Modify specific concessions offered or demanded\newline{}Change the primary stated objective of the event\newline{}Alter the nature or scope of any proposed agreement \\
    Context & Change the geographical location of diplomatic events \newline{}Modify symbolic gestures or protocols observed between parties \\
    \bottomrule
    \end{tabular}%
  \label{tab:counterfactual_categories}%
\end{table*}%

To explore how specific alterations can induce desired shifts in public sentiment, we develop an iterative counterfactual generation algorithm  \ref{alg:detailed_counterfactual}. This process begins by defining a target sentiment transformation for an original diplomatic event description (for instance, aiming to shift its predicted sentiment from `Negative' to `Neutral,' or from `Neutral' to `Positive'). Starting with the original text, our Large Language Model (LLM) first applies a modification from a single, predefined counterfactual category (e.g., `participants'). The sentiment of this newly generated counterfactual text is then predicted using our fine-tuned sentiment detection model. If this predicted sentiment aligns with the predefined target, the generation process for that specific transformation concludes, and the successful counterfactual is recorded. However, if the target sentiment is not achieved, this initial modification is retained, and the LLM proceeds to apply a new modification from the next counterfactual category (e.g., `process') to the already altered text. This sequential modification and evaluation cycle continues through our established list of counterfactual categories until the desired sentiment outcome is reached or all categories have been attempted for that event.

\begin{algorithm*}[htp]
\caption{Iterative Counterfactual Generation Algorithm}
\label{alg:detailed_counterfactual}
\begin{algorithmic}[1]
\Require $D_0$: Original diplomatic event description
\Require $S_{target}$: Target sentiment classification
\Require $C = \{participants, process, communication, substance, context\}$: Counterfactual categories
\Require $LLM$: Large Language Model for text modification
\Require $M_{sentiment}$: Fine-tuned sentiment detection model
\Ensure transformed text with target sentiment or failure indication

\Procedure{GenerateCounterfactual}{$D_0, S_{target}, C, LLM, M_{sentiment}$}
    \State $D_{current} \leftarrow D_0$
    \State $transformations \leftarrow []$ \Comment{Track applied modifications}
    
    \For{$category \in C$}
        \State \Comment{Apply modification from current category}
        \State $prompt \leftarrow$ ConstructPrompt($D_{current}$, $category$)
        \State $D_{modified} \leftarrow LLM.generate(prompt)$
        
        \State \Comment{Evaluate sentiment of modified text}
        \State $S_{current} \leftarrow M_{sentiment}.predict(D_{modified})$
        
        \State \Comment{Record transformation details}
        \State $transformation \leftarrow \{category, D_{current}, D_{modified}, S_{current}\}$
        \State $transformations.append(transformation)$
        
        \If{$S_{current} = S_{target}$}
            \State \textbf{return} $(D_{modified}, transformations, \textit{SUCCESS})$
        \EndIf
        
        \State \Comment{Update current text for next iteration}
        \State $D_{current} \leftarrow D_{modified}$
    \EndFor
    
    \State \textbf{return} $(D_{current}, transformations, \textit{FAILURE})$
\EndProcedure

\end{algorithmic}
\end{algorithm*}

\section{Experiments and results analysis}
\label{sec:experimentsandresultsanalysis}

\subsection{Data description}
In this study, we selected a set of common and representative diplomatic events for our analysis, as illustrated in Table \ref{tab:Examples of Event Data}. This data set comprises 21 distinct diplomatic events classified into 10 categories of diplomacy. It captures 8,000 Reddit posts related to diplomatic events, along with comments from users from various countries, backgrounds, communities, and groups. Sentiment is evaluated using a weighted average approach that incorporates both upvotes and downvotes.
The decision to categorize diplomatic events into this set of types—ranging from bilateral and summit diplomacy to more specialized forms like economic, humanitarian, and digital diplomacy—is a methodological choice based on the complex, multifaceted nature of modern international relations. Different diplomatic activities operate through distinct channels, involve different actors, communicate different messages, and ultimately, trigger different public reactions. 

A comprehensive list of categories is necessary to reflect the reality of 21st-century diplomacy. For example, 
For bilateral diplomacy, we selected high-stakes negotiations such as the US-China Strategic and Economic Dialogue, the sensitive India-Pakistan talks on the Indus Waters Treaty, and the economically crucial UK-EU post-Brexit trade negotiations.
In multilateral diplomacy, we included the globally vital United Nations Climate Change Conference and the WHO's response to pandemics, both of which address critical global challenges requiring broad international cooperation. For summit diplomacy, we chose major economic and political forums like the G20, BRICS, and APEC summits, which bring together world leaders to shape global governance and economic policy. These events were selected for their high impact, extensive media coverage, and ability to generate significant public discourse. Digital and science diplomacy represent the frontier of international relations. Digital diplomacy on platforms like Twitter allows for direct, unfiltered communication with global publics, making it a highly dynamic and often volatile domain for sentiment formation. Science diplomacy, involving collaboration on issues like climate change or pandemics, often generates sentiment related to global cooperation and shared progress.

Furthermore, analyzing these separately is crucial because the public reactions are fundamentally different for different types.
For example, economic and cultural diplomacy are now central pillars of soft power. A trade deal may be evaluated by the public based on perceived economic costs and benefits, while a cultural exchange might be judged on its ability to foster goodwill and a positive national image. 
Humanitarian, nuclear, and migration diplomacy are functionally specific and often highly contentious. Public sentiment towards humanitarian diplomacy can be driven by empathy and moral arguments, while reactions to nuclear diplomacy are often shaped by perceptions of security and existential threat.

This categorization enables us to identify the specific drivers of sentiment, generating far more actionable recommendations for policymakers and practitioners. A generic finding that ``positive tone improves sentiment" is of limited use. However, a more specific finding, such as ``a conciliatory tone is highly effective in humanitarian diplomacy but has little effect in nuclear diplomacy where perceived strength is more valued," provides concrete, context-specific guidance. It allows us to ask more sophisticated questions, for example, 
are the drivers of sentiment for summit diplomacy (often personality-driven and symbolic) different from those for Economic Diplomacy (often focused on material outcomes)?
Does the public react more strongly to the process in multilateral settings versus the substance in bilateral ones?
Is public sentiment more volatile in Digital Diplomacy than in traditional forms?
By treating each category as a distinct context, our framework can isolate variables and determine if the importance of a feature (like the tone of communication or the actors involved) is consistent across all types of diplomacy or if it is highly dependent on the specific context. This is crucial for building a model that is not only predictive but also explanatory.

\begin{table*}[htbp]
  \centering
  \normalsize
  \caption{Examples of Event Data}
    \begin{tabular}{|l|p{40.75em}|}
    \toprule
    Type  & Event \\
    \midrule
    \multicolumn{1}{|l|}{\multirow{3}[6]{*}{Bilateral Diplomacy}} & US-China Strategic and Economic Dialogue \\
\cmidrule{2-2}          & India-Pakistan talks on the Indus Waters Treaty \\
\cmidrule{2-2}          & UK-EU negotiations on post-Brexit trade arrangements \\
    \midrule
    \multirow{2}[4]{*}{Multilateral Diplomacy} & United Nations Climate Change conference \\
\cmidrule{2-2}          & WHOPandemic \\
    \midrule
    \multirow{3}[6]{*}{Summit Diplomacy} & G20 Summit \\
\cmidrule{2-2}          & BRICS Summit \\
\cmidrule{2-2}          & APECSummit \\
    \midrule
    \multirow{3}[6]{*}{Digital Diplomacy} & Foreign ministry using Twitter (X) to announce policy or condemn actions \\
\cmidrule{2-2}          & Embassies using Facebook for outreach to local populations \\
\cmidrule{2-2}          & Leaders engaging in "Zoom diplomacy" for virtual summits. \\
    \midrule
    \multirow{3}[6]{*}{Science Diplomacy} & International Thermonuclear Experimental Reactor (ITER) collaboration \\
\cmidrule{2-2}          & Global Polio Eradication Initiative involving multiple countries and organizations. \\
\cmidrule{2-2}          & US-Russia cooperation on the International Space Station (ISS). \\
    \midrule
    \multirow{3}[6]{*}{Economic Diplomacy} & China's Belt and Road Initiative (BRI) involving infrastructure investments.  \\
\cmidrule{2-2}          & US negotiating the Trans-Pacific Partnership (TPP) or its successor CPTPP \\
\cmidrule{2-2}          & EU imposing trade tariffs or negotiating free trade agreements (FTAs). \\
    \midrule
    \multirow{3}[6]{*}{Cultural Diplomacy} & Japan promoting its pop culture (anime, manga) globally ("Cool Japan"). \\
\cmidrule{2-2}          &  France organizing international art exhibitions (e.g., Louvre Abu Dhabi).  \\
\cmidrule{2-2}          & South Koreas K-Pop diplomacy. \\
    \midrule
    \multicolumn{1}{|l|}{\multirow{3}[6]{*}{Humanitarian Diplomacy}} & Red Cross/Red Crescent negotiating access to conflict zones.  \\
\cmidrule{2-2}          & UN agencies (like WFP or UNHCR) negotiating aid delivery in crisis areas. \\
\cmidrule{2-2}          & Diplomats advocating for the protection of civilians in armed conflict. \\
    \midrule
    \multirow{3}[6]{*}{Nuclear Diplomacy} & Negotiations for the New START Treaty between US and Russia.  \\
\cmidrule{2-2}          & Six-Party Talks concerning North Korea's nuclear program. \\
\cmidrule{2-2}          & P5+1 negotiations with Iran on its nuclear program (JCPOA). \\
    \midrule
    \multirow{3}[6]{*}{Migration Diplomacy} & EU-Turkey deal on managing refugee flows. \\
\cmidrule{2-2}          &  US negotiations with Mexico and Central American countries on border control. \\
\cmidrule{2-2}          & Germany's discussions on labour migration agreements with non-EU countries. \\
    \bottomrule
    \end{tabular}%
  \label{tab:Examples of Event Data}%
\end{table*}%

\subsection{Sentiment Analysis for Diplomatic Events}
We fine-tuned a language model that can predict the public sentiment, so we use it as a proxy for the public. We begin by 
analyzing the public sentiment of different groups for the same diplomatic event. The groups are divided into two key dimensions: geographical actors and thematic areas.
For geographical actors (China, India, the US, Non-US, Europe), this allows us to analyze how sentiment towards diplomatic events differs based on national or regional perspectives. It helps answer questions like, ``Is an event perceived more positively by users from Europe than from the US?" This is crucial for understanding the geopolitical landscape of public opinion. The ``Non-US" category provides a valuable point of comparison to isolate US-specific sentiment.
For thematic areas (Economics, Politics, Healthcare, Climate), this dimension allows us to analyze how sentiment is shaped by the core topic of the diplomatic event. It helps answer questions like, ``Are economic diplomatic events generally viewed more negatively than those related to climate?" This provides insight into which specific issue areas are most sensitive or successful in the public eye.
Analyzing how different groups react to the same diplomatic event is crucial because public sentiment is typically not monolithic. 
By breaking down the analysis by group, we can identify which specific publics are supportive, opposed, or indifferent to a diplomatic action. This provides a much richer and more actionable understanding. For policymakers and communicators, this granular insight is vital for tailoring messages, addressing specific concerns, and understanding the true political landscape surrounding a diplomatic event. 

\begin{figure*}[htp]
\caption{This table presents the descriptive statistics for sentiment scores related to key national/regional actors and thematic areas. The sentiment scores range from -1 (most negative) to +1 (most positive), with 0 indicating neutral sentiment.}
    \centering
    \includegraphics[width=1\linewidth]{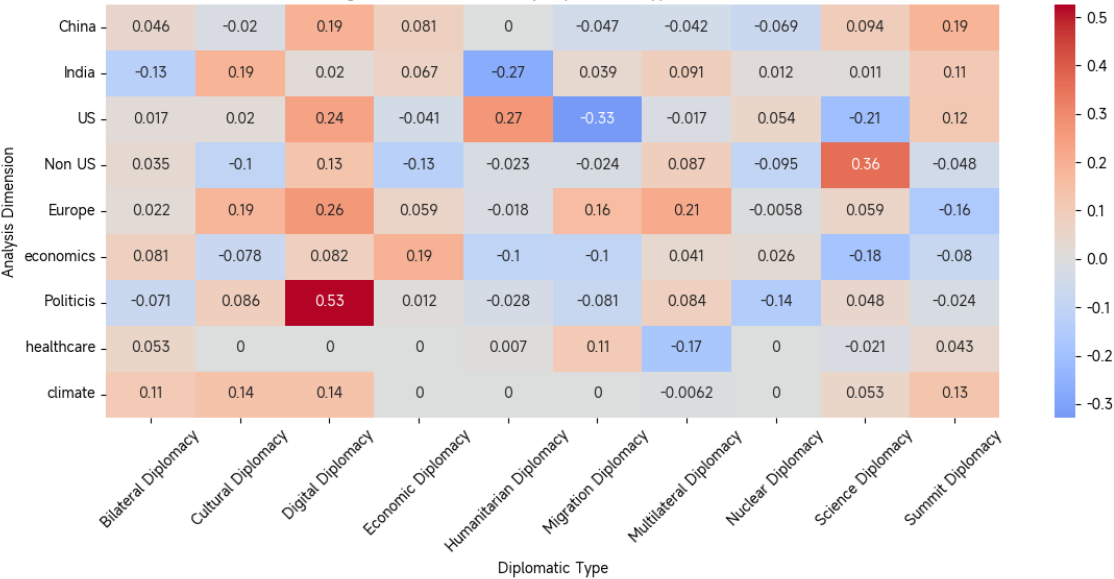}
    \label{fig:sentimentanalysis}
\end{figure*}

We present the sentiment results for diplomatic events in Figure \ref{fig:sentimentanalysis}. This heatmap reveals several significant trends and specific points of high sentiment, indicating that public reaction to diplomacy is highly dependent on both the actors involved and the nature of the event. 
The most striking finding is the exceptionally high positive sentiment score (0.53) at the intersection of Politics and Digital Diplomacy. This suggests that when political topics are handled through digital diplomatic channels, the public reaction is strongly favorable. This could indicate an appreciation for the directness, transparency, or modern approach of digital statecraft in the political realm.
Cultural diplomacy consistently elicits positive or near-positive sentiment from multiple actors, including India (0.19) and Europe (0.19). This suggests that cultural exchanges and soft power initiatives are generally well-received by the public.
The Economics dimension shows a pattern of moderately negative to positive sentiment. For instance, it is viewed positively in the context of economic diplomacy (0.19) but negatively in humanitarian diplomacy (-0.10) and multilateral diplomacy (-0.10). This suggests that economic discussions are sensitive and their public reception depends heavily on the diplomatic framework in which they are presented.
The climate theme consistently registers neutral to positive sentiment across all diplomatic types where it is measured. This suggests that diplomatic efforts related to climate change are generally perceived as a positive or necessary form of international cooperation.

\subsection{Using Counterfactual Algorithm for Interventions to Public Sentiment}
In this experiment, we input the narratives of diplomatic events in algorithm \ref{alg:detailed_counterfactual}, and then start to intervene in public sentiments using pre-defined categories one by one. Recall that the algorithm will start modifying the original text from the first category. If the public sentiment is changed to neutral or positive sentiment, the algorithm succeeds. If not, the algorithm will continue adjusting the next step.
Table \ref{tab:category_breakdown} displays the results of the success rate of modifying public sentiment using the counterfactual algorithm.

\begin{table}[htbp]
\centering
\caption{Success Breakdown by Category}
\label{tab:category_breakdown}
\begin{tabular}{l r r r}
\hline
\textbf{Category} & \textbf{Count} & \textbf{\% of Successes} & \textbf{\% of Total} \\
\hline
Participants & 577 & 30.19\% & 20.91\% \\
Context & 547 & 28.62\% & 19.82\% \\
Communication & 352 & 18.42\% & 12.75\% \\
Substance & 234 & 12.24\% & 8.48\% \\
Process & 201 & 10.52\% & 7.28\% \\
\hline
\textbf{Total Successes} & \textbf{1,911} & \textbf{100\%} & \textbf{70\%} \\
\hline
\end{tabular}
\end{table}

The counterfactual algorithm demonstrated substantial effectiveness in transforming negative public sentiment to neutral or positive outcomes, achieving a 70\% success rate across 2,760 analysed articles. 
Additionally, 77.23\% of cases of public sentiment are modified successfully only using the first three steps (Participants, Context, and Communication).
This sequential intervention approach revealed distinct patterns in the types of modifications required for successful sentiment transformation. Among the successful interventions, combining modifications to participants (30.19\%) and context (28.62\%) proved most effective, collectively accounting for nearly 60\% of all successful transformations. This suggests that altering the actors involved in news narratives or reframing the situational context represents the most powerful levers for sentiment change. 

The 30.76\% of articles that remained unchanged despite sequential interventions may represent cases where negative sentiment is inherently resistant to counterfactual modification, potentially due to factual constraints or deeply embedded narrative structures that cannot be altered without compromising authenticity or factual accuracy.



\subsection{How The Event Factors Drive Public Sentiment (Ablation Study)}
In the previous experiment, we showed the performance of using the counterfactual algorithm. In this section, we want to demonstrate the performance for each single modification to validate the contribution of each category modification.
Table \ref{tab:success_rate_analysis} demonstrates varying success rates across different intervention categories in transforming negative diplomatic sentiment to neutral/positive states. Table \ref{tab:detailed_modifications} shows the details of the success rate for steps within each Category.
For example, the success rate of changing participants alone reached 9.92\%.

\begin{table}[htbp]
  \centering
  \caption{Success Rate Analysis for Negative to Neutral/Positive Changes}
  \label{tab:success_rate_analysis}
    \begin{tabular}{lccc}
    \toprule
    Category & Total Cases & Successful & Success Rate \\
    \midrule
    Participants & 8,466 & 840 & 9.92\% \\
    Process & 8,466 & 892 & 10.54\% \\
    Communication & 8,466 & 1,565 & 18.49\% \\
    Substance & 8,466 & 1,444 & 17.06\% \\
    Context & 5,644 & 958 & 16.97\% \\
    \bottomrule
    \end{tabular}
\end{table}

\begin{table*}[htbp]
  \centering
  \caption{Detailed Success Rate Analysis by Specific Modification Type}
  \label{tab:detailed_modifications}
  \begin{tabular}{p{8cm}ccc}
    \toprule
    Modification Type & Total & Successful & Success Rate \\
    \midrule
    \multicolumn{4}{l}{\textbf{Category: Participants}} \\
    \quad Replace the lead negotiator with alternative & 2,822 & 336 & 11.91\% \\
    \quad Include additional stakeholders & 2,822 & 330 & 11.69\% \\
    \quad Exclude certain parties from talks & 2,822 & 174 & 6.17\% \\
    \midrule
    \multicolumn{4}{l}{\textbf{Category: Process}} \\
    \quad Change the negotiation format & 2,822 & 206 & 7.30\% \\
    \quad Modify the timing of diplomatic initiatives & 2,822 & 269 & 9.53\% \\
    \quad Alter the use of coercive measures & 2,822 & 417 & 14.78\% \\
    \midrule
    \multicolumn{4}{l}{\textbf{Category: Communication}} \\
    \quad Modify the tone of official statements & 2,822 & 702 & 24.88\% \\
    \quad Change the level of publicity & 2,822 & 276 & 9.78\% \\
    \quad Reframe key issues in different terms & 2,822 & 587 & 20.80\% \\
    \midrule
    \multicolumn{4}{l}{\textbf{Category: Substance}} \\
    \quad Modify specific concessions offered & 2,822 & 298 & 10.56\% \\
    \quad Change the primary objective & 2,822 & 610 & 21.62\% \\
    \quad Alter the nature of any agreement & 2,822 & 536 & 18.99\% \\
    \midrule
    \multicolumn{4}{l}{\textbf{Category: Context}} \\
    \quad Change the location of diplomatic events & 2,822 & 278 & 9.85\% \\
    \quad Modify symbolic gestures between parties & 2,822 & 680 & 24.10\% \\
    \bottomrule
  \end{tabular}
\end{table*}

The results show that Communication-related modifications demonstrated the highest average success rate (18.5\%). 
The superior performance of communication interventions aligns with discourse analysis theories emphasizing linguistic mediation in diplomatic sentiment transformation. The 8.57 percentage point gap between communication and process interventions suggests that how messages are framed matters more than procedural adjustments.
The comparable success rates of substance (17.06\%) and context (16.97\%) interventions support the constitutive model of diplomatic communication, where substantive content and situational framing exhibit interdependent effects on sentiment modulation.
The unexpectedly low success rate of direct participant interventions (9.92\%) challenges conventional actor-centric diplomatic models, potentially indicating that LLM-mediated sentiment change operates more effectively through discursive rather than personal channels.

Notably, Communication, Substance, and Context strategies exceeded 20\% success rates. The least effective strategies involved excluding parties from talks (6.2\%) and changing negotiation formats (7.0\%), suggesting structural changes face greater implementation barriers.
The 24.9\% success rate for tone modification aligns with existing literature on framing effects in political communication. The underperformance of participant-related changes challenges conventional diplomatic wisdom about personnel adjustments, potentially indicating institutional inertia in negotiation contexts.
The success gradient from communicative to structural modifications suggests an inverse relationship between intervention depth and effectiveness - surface-level linguistic adaptations prove more implementable than fundamental process changes.

That is to indicate, government civil servants, journalists, and others can use this framework to judge the public sentiment trend of events and make precise predictions. In the events that have already occurred, if one wants to adjust the public opinion consequences brought about by this matter, such as conducting timely crisis public relations and public opinion guidance, etc., this framework can be used for adjustment and prediction to achieve a positive public opinion trend.

\section{Conclusion}
\label{sec:conclusion}

The primary objective of this study was to address a significant gap in understanding public sentiment toward diplomatic events—a critical factor shaping international relations and policy outcomes. Traditional methods, such as surveys and expert assessments, have struggled to capture the nuanced and dynamic nature of public opinion in real-time. To bridge this gap, we introduced a novel computational framework combining fine-tuned language models with counterfactual analysis to systematically identify the drivers of public sentiment in diplomatic contexts.
Our approach involved three core stages: (1) constructing a specialized dataset by pairing news articles about diverse diplomatic events with sentiment labels derived from global Reddit discussions; (2) fine-tuning a BERT-based model to predict public sentiment from these event narratives, adapting it to the intricacies of diplomatic discourse; and (3) employing a Large Language Model (LLM) to generate counterfactual versions of event descriptions by altering predefined features across categories such as participants, process, communication style, substance, and context. By analysing how these controlled modifications influenced sentiment predictions, we aimed to uncover which event characteristics most significantly shaped public perception.

Our findings revealed that systematic alterations to specific features of diplomatic events could shift public sentiment from negative to neutral or positive, with a 70\% success rate.  Changes rooted in Framing Theory, such as modifying communication style and contextual details, were identified as particularly influential factors. For instance, changes in communication style, framing of substantive outcomes, and contextual details were identified as particularly influential factors. These results highlight the potential for strategic interventions in how diplomatic events are framed and communicated to the public. Furthermore, our framework demonstrated the utility of explainable sentiment analysis in computational social science, offering deeper insights into the complex interplay between diplomatic actions and public opinion.  

This research contributes to multiple domains. First, it advances new computational methods for analysing geopolitical discourse, providing a new tool for policymakers and researchers to gauge public sentiment more effectively. Second, it underscores the importance of transparency in AI-driven analyses, particularly in sensitive areas like diplomacy, where accountability and actionable insights are paramount. Finally, by identifying key drivers of public sentiment, our framework can inform strategies for digital diplomacy and nation branding, enabling states to craft narratives that resonate positively with global audiences.

Future work could expand the scope to include multilingual datasets, and cross-cultural comparisons would provide a more comprehensive understanding of global public sentiment dynamics.

In conclusion, this study provides a robust foundation for exploring the intricate relationship between diplomatic events and public sentiment. By leveraging advanced computational techniques, we offer a methodological advancement with practical implications for both academic research and real-world policy-making. As digital platforms continue to shape political discourse, frameworks like ours will be essential for navigating the evolving landscape of international relations.


\bibliographystyle{apacite}  
\bibliography{references}

\end{document}